\documentclass{article}

\usepackage{graphicx} 

\usepackage[final]{neurips_2024}
\usepackage{hyperref}
\hypersetup{colorlinks=true,linkcolor=black,citecolor=black,anchorcolor=black}
\usepackage[utf8]{inputenc} 
\usepackage[T1]{fontenc}    
\usepackage{hyperref}       
\usepackage{url}            
\usepackage{booktabs}       
\usepackage{amsfonts}       
\usepackage{nicefrac}       
\usepackage{xcolor}         

\title{Oracle Bone Inscriptions Multi-modal Dataset}

\author{%
   Bang Li$^{1*}$, Donghao Luo$^{2*}$, Yujie Liang$^{3}$, Jing Yang$^{1}$, Zengmao Ding$^{1}$, Xu Peng$^{2}$, Boyuan Jiang$^{2}$,\\
   Shengwei Han$^{1}$, Dan Sui$^{1}$, Peichao Qin$^{5}$, Pian Wu$^{4}$, Chaoyang Wang$^{4}$, Yun Qi$^{2}$, Taisong Jin$^{3}$,  \\
   Chengjie Wang$^{2}$, Xiaoming Huang$^{2}$, Zhan Shu$^{4}$, Rongrong Ji$^{3}$, Yongge Liu$^{1}$, Yunsheng Wu$^{2}$ \\
  \\
    $^1$Key Laboratory of Oracle Bone Inscriptions Information Processing,\\
      Ministry of Education of China, Anyang Normal University \\
     $^2$Youtu Lab, Tencent \\
    $^3$Key Laboratory of Multimedia Trusted  Perception and Effcient Computing,\\
        Ministry of Education of China, Xiamen University \\
    $^4$Digital Culture Lab,  Tencent SSV (Sustainable Social Value Dept) \\
   $^5$University of Cambridge
}

\begin{document}

\maketitle

\begin{abstract}
Oracle bone inscriptions(OBI) is the earliest developed writing system in China, bearing invaluable written exemplifications of early Shang history and paleography. However, the task of deciphering OBI, in the current climate of the scholarship, can prove extremely challenging. Out of the 4,500 oracle bone characters excavated, only a third have been successfully identified. Therefore, leveraging the advantages of advanced AI technology to assist in the decipherment of OBI is a highly essential research topic. However, fully utilizing AI's capabilities in these matters is reliant on having a comprehensive and high-quality annotated OBI dataset at hand whereas most existing datasets are only annotated in just a single or a few dimensions, limiting the value of their potential application. For instance, the Oracle-MNIST dataset only offers 30k images classified into 10 categories.
Therefore, this paper proposes an Oracle Bone Inscriptions Multi-modal Dataset(OBIMD), which includes annotation information for 10,077 pieces of oracle bones. Each piece has two modalities: pixel-level aligned rubbings and facsimiles. The dataset annotates the detection boxes, character categories, transcriptions, corresponding inscription groups, and reading sequences in the groups of each oracle bone character, providing a comprehensive and high-quality level of annotations. This dataset can be used for a variety of AI-related research tasks relevant to the field of OBI, such as OBI Character Detection and Recognition, Rubbing Denoising, Character Matching, Character Generation, Reading Sequence Prediction, Missing Characters Completion task and so on. We believe that the creation and publication of a dataset like this will help significantly advance the application of AI algorithms in the field of OBI research.



\end{abstract}

\section{Introduction}
As the earliest developed Chinese writing system, oracle bone inscriptions (OBI) possess significant research value.
However, the application of Artificial Intelligence(AI) in this field is limited due to the lack of relevant data.
The existing datasets usually offer simple annotation information. \cite{liu2020spatial} is confined to detecting OBI on rubbings without the capability for recognition, and \cite{huang2019obc306,guo2015building,han2020isobs,fu2022improvement,yue2022dynamic,fujikawa2022recognition} are limited to recognizing only a small subset of OBI.

To address these limitations, we propose an Oracle Bone Inscriptions Multi-modal Dataset (OBIMD), the world's first comprehensive dataset encompassing multi-view information for OBI research. The dataset employs oracle bone rubbings as input, supplemented with precisely aligned facsimiles, transcriptions, and annotated single-character images, thereby providing a holistic translation of the content.

The OBI research can divided into multiple machine learning tasks, including OBI Detection and Recognition, Rubbing Denoising, Character Matching, Character Generation, Reading Sequence Prediction, Missing Characters Completion and so on.
These machine learning tasks will support key steps in OBI translation, such as character recommendation, hypothesis verification, and evidence collection, significantly advancing OBI research through AI technology.

\section{Dataset Statistics}

\begin{figure*}[h]
    \centering
\includegraphics[width=1.0\linewidth]{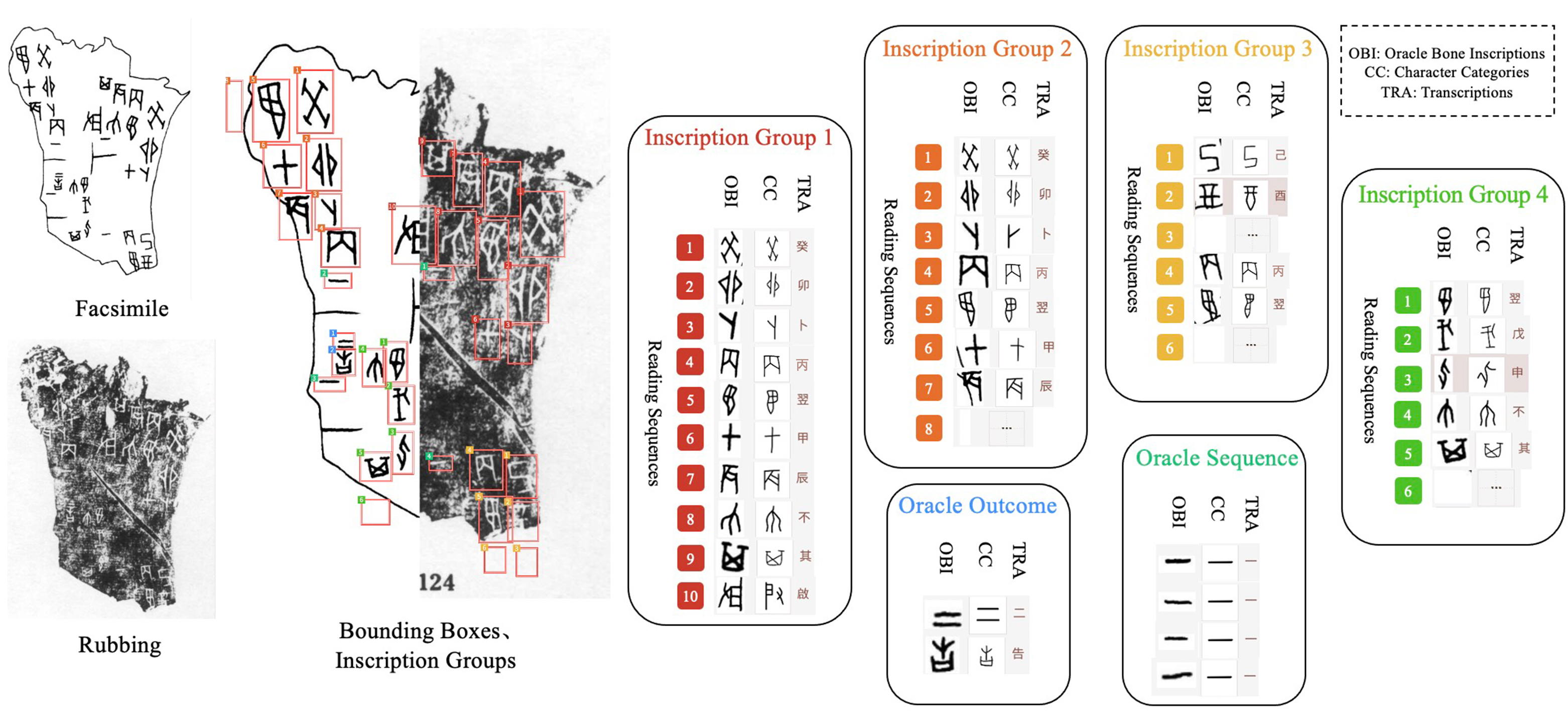}
    \vspace{-4mm}
    \caption{Data Sample.}
    \label{fig:arch0}
    \vspace{-1em}
\end{figure*}

\subsection{Dataset Composition}
The OBIMD dataset rubbings primarily from the "Oracle Bone Inscriptions Collection\cite{Guo1978OracleBoneInscriptionsCollection}". We performed uniform sampling based on the number of rubbings from five historical periods of the Shang Dynasty, randomly selecting 10,077 rubbings proportionally. Additionally, we included 164 rubbings from the "Yinxu Huayuanzhuang East Oracle Bones\cite{huadong}" to ensure the dataset's historical period distribution accurately reflects the actual distribution of oracle bones.
The dataset annotations are divided into three parts:
\begin{itemize}
\item[$\bullet$]

Digital Facsimiles: These are pixel-by-pixel correspondences to the rubbings, created by oracle bone scholars who manually transcribe the characters and edges from the rubbings using comprehensive knowledge and various sources. The facsimiles are selected from the "Comprehensive Series of Oracle Bone Facsimiles\cite{Mobendaxi}" and undergo pre-processing steps such as character detection, matching, and registration, followed by manual verification to ensure accuracy.

\item[$\bullet$]
Text Content Annotations: This includes the detection boxes, character categories, transcriptions, corresponding inscription groups, and reading sequences in the groups of each oracle bone character. The character bounding boxes identify the location of characters on the rubbings. Character categories are based on the "Oracular Digital Platform" character library (https://oracular.azurewebsites.net/glyphs), which uses a two-level category structure. Attributes mask special conditions like contentious or missing parts. Inscription groups help in identifying characters that belong to the same sentence and categorizing sentence attributes into inscription sentence, oracle outcome, oracle sequence, and uncertain. The dataset contains 115,319 bounding boxes (including placeholders), with 93,652 characters and 21,667 placeholders forming 21,941 inscription groups, 3,785 oracle sequences, and 407 oracle outcomes.
\item[$\bullet$]
Character Information: This part includes images of sub-characters, and mapping tables between sub-characters and main characters, as well as between sub-characters and transcriptions. These materials support tasks such as OBI forms matching and characters generating. Each OBI in the "Oracular Digital Platform" character library has a unique UID, annotated with a corresponding character encoding sourced from the authoritative website Mebag (https://www.mebag.com/index/).
\end{itemize}

\begin{figure*}[h]
    \centering
\includegraphics[width=0.3\linewidth]{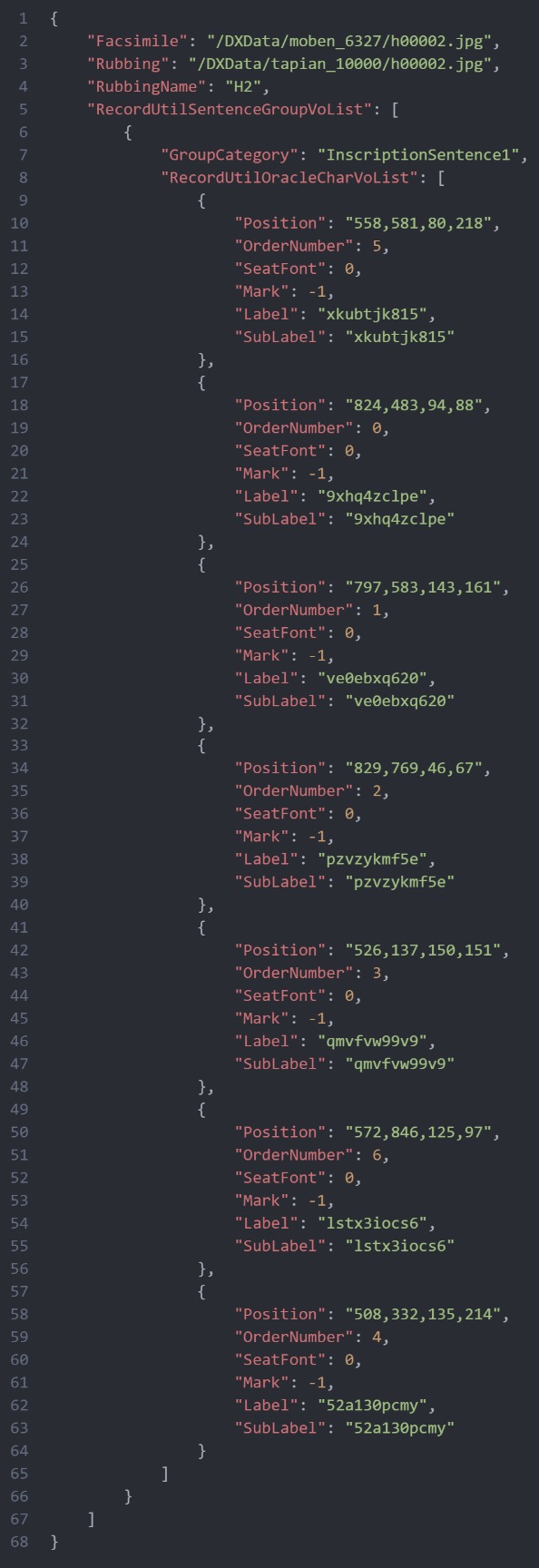}
    \vspace{2mm}
    \caption{Data Annotation.}
    \label{fig:arch1}
    \vspace{-1em}
\end{figure*}

\subsection{Dataset Annotation}
As shown in Figure~\ref{fig:arch1}, Dataset annotation is as follows:
\begin{itemize}
\item[$\bullet$]
Facsimile: [String Type] Path to the complete facsimile image.
\item[$\bullet$]
Rubbing: [String Type] Path to the complete rubbing image.
\item[$\bullet$]
RubbingName: [String Type] The current name of the rubbing.
\item[$\bullet$]
RecordUtilSentenceGroupVoList: [Array Type] List of inscription groups contained in the rubbing.
\item[$\bullet$]
GroupCategory: [String Type] Classification of inscription groups. Categories include "inscription sentence", "oracle sequence", "oracle outcome", and "uncertain".
\item[$\bullet$]
RecordUtilOracleCharVoList: [Array Type] List of the characters belonging to the inscription group.
\item[$\bullet$]
Position: [String Type] Bounding box position, labeled by four integers representing the top-left corner coordinates and dimensions (width and height) in pixels, indicating the annotation position/current single character's position in the complete rubbing or facsimile.
\item[$\bullet$]
OrderNumber: [Integer Type] Marks the sequence of the bounding box within the inscription group, starting from 0 and counting upwards, indicating the reading order of the current character within the inscription group.
\item[$\bullet$]
SeatFont: [Integer Type] Bounding box attribute label, indicating whether the bounding box is a placeholder for a missing oracle bone character. Default is marked as 0, where 0 means no placeholder; 1 indicates a placeholder, representing a missing OBI with only a bounding box and oracle sequence but no corresponding character image.
\item[$\bullet$]
Mark: [Integer Type] Bounding box attribute label, using numbers to identify special cases. Default is marked as -1, representing normal conditions. Special (contentious) cases include: 0 for damaged characters that cannot be accurately identified; 1 for disputed attribution of the oracle bone character's category; 2 for discrepancies between the character's rubbing and facsimile information, existing only on the rubbing, with doubts whether the rubbing image should be considered an oracle bone character; 3 for discrepancies between the character's rubbing and facsimile information, existing only on the facsimile, where the facsimile character form is not derived from the rubbing image.
\item[$\bullet$]
Label: [String Type] Character category label, representing the UID of the category to which the character belongs, where the character represents one of the types of OBI.
\item[$\bullet$]
SubLabel: [String Type] Character category label, representing the UID of the sub-category to which the character belongs, where the sub-character represents different structures of the same type of OBI.
\end{itemize}
\section{Methods}

The OBIMD dataset’s annotations are divided into three sections. Digital facsimiles undergo alignment via a previously developed character-based registration algorithm, complemented by manual verification and adjustments. The modern Chinese character mapping table is sourced from the "Oracular Digital Platform" character library, requiring no specialized knowledge. However, annotating the text content on oracle bone rubbings demands high expertise in recognizing and locating characters within the character library, limiting the availability of qualified annotators. To address this, we implemented a barrier-free initial tagging process supplemented by expert review, supported by a specially developed online annotation tool.

The platform facilitates pixel-by-pixel comparisons between rubbings and facsimiles, providing original texts and transcriptions for each piece. Initial annotators can identify character positions and perform preliminary sentence segmentation with the aid of facsimiles and published transcription books\cite{moshiquanbian}. Each character is also matched against the top 10 forms from the glyph library using a shape matching algorithm, and the tools for character handwriting and transcription are also provided to assist the task of accurate character identification. Additionally, an object detection algorithm is employed to pre-mark the bounding boxes on each rubbing to provide effective visual aid, enabling non-experts to complete near 80\% of the tasks, thus increasing annotating efficiency by approximately 60 times.

Post-initial annotation, oracle bone experts and trained graduate students in oracle bone information processing review the work. These students, highly educated in both the subjects of artificial intelligence and oracle bone script, are able to identify and correct any inaccuracies or inappropriate methods used during the initial batch of annotations. Under the expert guidance, they can also verify and refine the annotations. This results in a comprehensive multi-modal dataset of oracle bone inscriptions with detailed annotations.

\section{Conclusion}
To advance the application of AI technology in the field of oracle bone studies, this paper proposes a multi-modal oracle bone script dataset, which includes 10,077 pairs of thoroughly annotated rubbings and facsimiles. The rubbings and facsimiles are aligned on a pixel-level. For each oracle bone fragment, we have identified its detection boxes, character positions, as well as the corresponding glyph transcriptions, inscription groups, and their correct reading sequences. The publication of a comprehensive and high-quality annotation dataset such as this can provide detailed information and exhaustive training materials for each oracle bone fragment involved, enabling a wide range of AI-related tasks in the oracle bone scholarship.

\bibliographystyle{splncs04}
\bibliography{ref}
\end{document}